\documentclass[letterpaper, 10 pt, conference]{ieeeconf}  

\IEEEoverridecommandlockouts                              
\usepackage{graphicx}
\usepackage{stfloats}   
\usepackage{placeins}   
\usepackage{tabularx}
\usepackage{balance}
\usepackage[table,xcdraw]{xcolor}
\title{\LARGE \bf

 LIT-GS: LiDAR-Inertial-Thermal Gaussian Splatting for Illumination-Robust Mapping}

\usepackage{float}
\usepackage{bm}
\usepackage[utf8]{inputenc}
\usepackage[noend]{algpseudocode}
\usepackage{amsmath}
\usepackage{amssymb}    
\usepackage[utf8]{inputenc}
\usepackage{xcolor}
\usepackage{graphicx}
\usepackage{hyperref}
\usepackage{booktabs}
\usepackage{color}
\usepackage{array}
\usepackage{booktabs}
\usepackage{multirow}
\usepackage{makecell}
\usepackage[space]{cite}
\usepackage[linesnumbered,boxed,ruled,commentsnumbered]{algorithm2e}
\graphicspath{{figures/}}

\begin{document}

\author{
Shikuan Shi$^{1,*}$, Chunran Zheng$^{2,*}$, Jiaming Xu$^{1}$, Tianyong Ye$^{1}$, Tao Yu$^{1}$, Yukang Cui$^{1\dagger}$
\thanks{* Equal contribution. $^{\dagger}$ Corresponding Author.}
\thanks{This work was partially supported by National Natural Science Foundation of China (61903258), Guangdong Basic and Applied Basic Research Foundation (2024A1515030153), and the Project of Department of Education of Guangdong Province (2023ZDZX4046).}
\thanks{$^{1}$Shikuan Shi, Jiaming Xu, Tianyong Ye, Tao Yu and Yukang Cui are with the College of Mechatronics and Control Engineering, Shenzhen University, Shenzhen 518060, China. (e-mail: cuiyukang@gmail.com)}
\thanks{$^{2}$Chunran Zheng is with the Department of Mechanical Engineering, The University of Hong Kong, Hong Kong SAR, China. (e-mail: zhengcr@connect.hku.hk)}
\thanks{This paper has been accepted by the IEEE/RSJ International Conference on Intelligent Robots and Systems (IROS), 2026.}
\thanks{© 2026 IEEE. Personal use of this material is permitted. For all other uses, permission from IEEE is required.}
}
\maketitle

\begin{abstract}

Gaussian Splatting has enabled real-time neural rendering, yet existing LiDAR–inertial–visual (LIV) Gaussian mapping pipelines remain fragile under illumination changes and texture-deficient scenes due to their reliance on RGB photometric cues. We present LIT-GS, a LiDAR–inertial–thermal Gaussian Splatting framework that injects LiDAR-derived plane geometry as an explicit constraint in both pose/structure refinement and Gaussian optimization. Specifically, we exploit LIV visual map points as confidence-aware cross-modal anchors to establish reliable thermal–LiDAR associations, and incorporate weighted LiDAR point-to-plane residuals into bundle adjustment to jointly refine camera poses and 3D points under weak thermal supervision. Building on the refined structure, we further introduce a LiDAR-plane-regularized differentiable splatting objective that constrains rendered 3D points to align with locally observed planes, mitigating surface thickening and structural drift in low-contrast thermal imagery. Experiments on proprietary sequences and public datasets demonstrate that LIT-GS consistently improves geometric accuracy and rendering quality over state-of-the-art LIV-based Gaussian Splatting baselines, particularly in challenging lighting conditions.

\end{abstract}

\section{INTRODUCTION}

Recent advances in 3D Gaussian Splatting (3DGS) have enabled real-time neural rendering with high fidelity and efficiency \cite{kerbl20233d}. 
By representing a scene as a set of anisotropic Gaussian primitives optimized via differentiable splatting, 3DGS offers an attractive speed-quality trade-off compared with NeRF-style radiance fields \cite{mildenhall2021nerf}. 
This efficiency has rapidly motivated radiance-field mapping for robotics and large-scale environments, yet most existing 3DGS-based mapping pipelines still depend heavily on RGB imagery and Structure-from-Motion (SfM) to initialize camera poses and sparse geometry \cite{schonberger2016structure}.

Reliance on visible imagery poses a fundamental limitation. 
Under illumination changes or texture-deficient scenes, photometric cues become unstable, degrading correspondence quality and pose estimation \cite{xu2026fast}.
Although LiDAR provides metric geometry, existing LiDAR-inertial-visual Gaussian Splatting systems mainly use LiDAR for initialization or weak regularization, while the core optimization remains dominated by RGB photometric supervision \cite{hong2024liv,hong2025gs,xiao2024liv}.

To address these challenges, this paper proposes LIT-GS, a LiDAR-inertial-thermal Gaussian Splatting framework that enforces LiDAR geometry as an explicit and persistent constraint in both pose/structure refinement and Gaussian optimization. 
Unlike RGB cameras, thermal imaging captures emitted thermal radiation and is largely insensitive to visible illumination, providing more stable supervision under extreme lighting variations. 
Recent thermal 3DGS studies further demonstrate the feasibility of optimizing radiance fields using thermal observations \cite{chen2024thermal3d,lu2024thermalgaussian}. 
However, illumination-robust supervision alone is insufficient to guarantee metric-consistent Gaussian mapping. 
Existing thermal 3DGS methods primarily target novel-view synthesis and often lack a tightly coupled pose/structure refinement loop \cite{chen2024thermal3d,lu2024thermalgaussian}, 
whereas current LIV Gaussian mapping pipelines remain largely RGB-driven and rarely enforce LiDAR geometry as a persistent constraint during Gaussian parameter learning \cite{hong2024liv,xie2024gs}. 
LIT-GS bridges this gap by (i) exploiting uncertainty-tagged visual map points from a modern LIV estimator as cross-modal geometric anchors \cite{zheng2022fast, zheng2024fast}, (ii) integrating weighted LiDAR point-to-plane residuals into bundle adjustment to jointly refine poses and triangulated structure, and (iii) introducing a LiDAR-plane-regularized splatting objective that constrains rendered 3D points toward locally observed LiDAR planes during Gaussian training. 
This two-level geometric injection, at both refinement and learning stages, suppresses surface-thickening artifacts and structural drift under weak thermal gradients.

\begin{figure}[t]
    \begin{center}
        {\includegraphics[width=1.0\columnwidth]{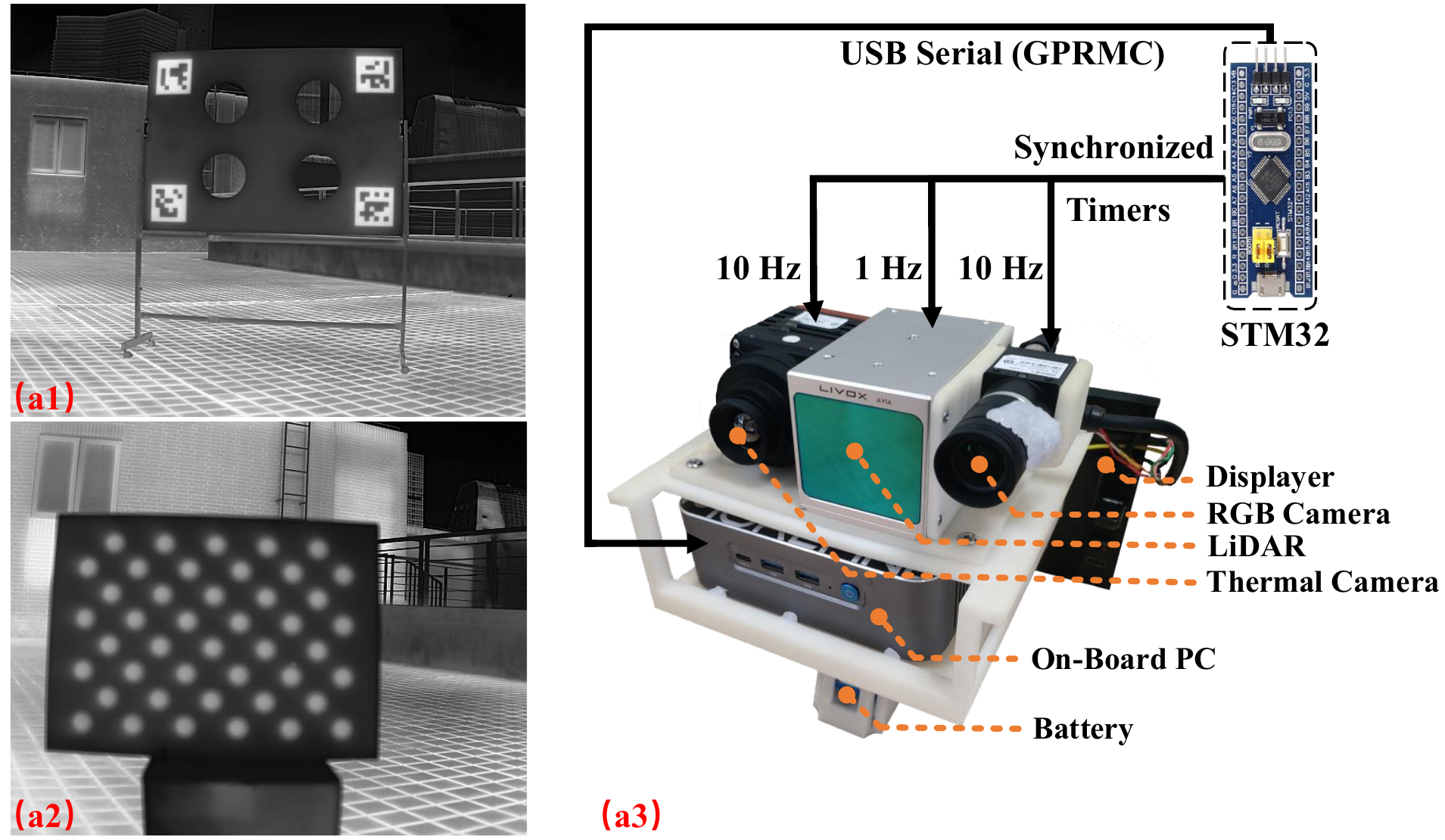}}
    \end{center}
    \vspace{-0.4cm}
    \caption{Hardware setup of LIT-GS. (a1) thermal observations of the FAST-Calib target under strong sunlight; (a2) thermal observations of the camera's internal reference target; (a3) device layout and the PPS-based time-synchronization scheme.}
    \label{fig:eqiup} 
    \vspace{-0.5cm}
\end{figure}

The main contributions are summarized as follows:
\begin{itemize}
    \item We propose LIT-GS, a LiDAR-inertial-thermal Gaussian Splatting framework that couples LiDAR plane constraints into both pose/structure refinement and Gaussian optimization, improving metric consistency under illumination changes and texture-deficient scenes.
    \item To stabilize thermal-LiDAR alignment, we develop a confidence-aware cross-modal anchoring and refinement scheme that uses uncertainty-annotated LIV map points as geometric anchors, fuses learned thermal correspondences, and integrates motion-adaptive weighting with LiDAR point-to-plane residuals into bundle adjustment.
    \item We further introduce a LiDAR-plane-regularized differentiable splatting objective with adaptive Gaussian map control, jointly enforcing thermal photometric fidelity and point-to-plane geometry to suppress surface thickening and structural drift, and validate the approach on both private and public datasets.
\end{itemize}

\begin{figure*}[t]
    \begin{center}
        {\includegraphics[width=2.05\columnwidth]{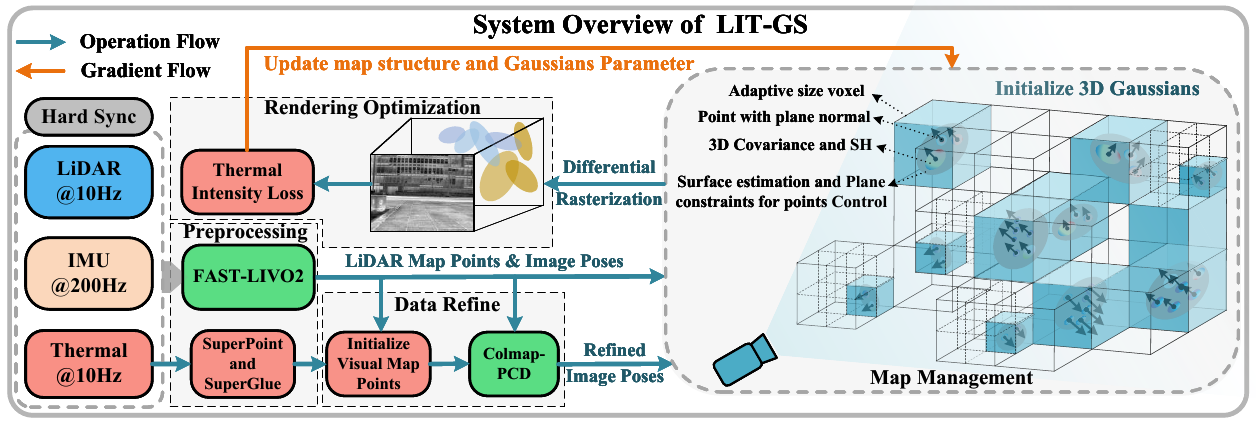}}
    \end{center}
    \vspace{-0.5cm}
    \caption{System overview of LIT-GS.}
    \label{fig:system_overview} 
    \vspace{-0.5cm}
\end{figure*}

\section{RELATED WORKS}

3DGS enables efficient radiance-field rendering by optimizing anisotropic Gaussian primitives via differentiable splatting \cite{kerbl20233d}. Building on this representation, recent work has extended 3DGS to challenging sensing conditions and robotics-oriented mapping. 
Moreover, 3D Gaussian primitives have been used as explicit map representations for online dense SLAM, as exemplified by SplaTAM in RGB-D settings \cite{keetha2024splatam}.
Thermal and multimodal 3DGS have been explored to improve robustness under illumination changes \cite{chen2024thermal3d,lu2024thermalgaussian}, while LiDAR-assisted Gaussian mapping introduces metric geometry for large-scale reconstruction \cite{hong2024liv,xie2024gs,xiao2024liv,liu2025gs}. Learned feature matchers such as SuperPoint and SuperGlue further improve correspondence reliability in low-contrast scenes \cite{detone2018superpoint,sarlin2020superglue}.

\subsection{Pure Thermal or Multimodal Gaussian Splatting}
For thermal novel-view synthesis, Chen \textit{et al.} \cite{chen2024thermal3d} incorporate physics-inspired priors into Gaussian rendering to suppress floaters and reduce boundary blurring. 
For multimodal fusion, Lu \textit{et al.} \cite{lu2024thermalgaussian} propose ThermalGaussian to learn a unified multimodal Gaussian representation with thermal-specific regularization. 
These studies show that thermal observations can provide effective supervision for 3DGS under illumination changes. 
However, they primarily target novel-view synthesis or multimodal rendering and typically do not address metric-consistent mapping under fast motion and weak texture, where tightly coupled pose/structure refinement and explicit geometric regularization become critical.

\subsection{Feature Extraction and Matching}
Classical SfM pipelines rely on hand-crafted detectors and descriptors and often degrade in weak-texture scenes or under severe appearance changes. Learning-based methods improve robustness by jointly learning detection and description, as exemplified by D2-Net \cite{dusmanu2019d2}. SuperPoint \cite{detone2018superpoint} and SuperGlue \cite{sarlin2020superglue} further enhance repeatability and matching reliability via attention-based context aggregation, yielding more stable correspondences under low contrast. This is particularly beneficial for thermal imagery, where gradients are weak and textures are sparse, thereby improving track connectivity. Nevertheless, correspondences alone are often insufficient under weak thermal gradients, and uncertainty-aware anchoring and geometric regularization remain essential to prevent drift and stabilize estimation.

\subsection{LIV Gaussian Splatting}
To bring 3DGS into robotics, recent systems integrate LiDAR geometry with visual(-inertial) estimation to bootstrap and optimize Gaussian maps for online mapping and rendering. LIV-GaussMap \cite{hong2024liv} initializes Gaussians from a LiDAR-inertial map and refines them with multi-view RGB photometric supervision. GS-LIVM \cite{xie2024gs} targets large-scale outdoor mapping, improving completeness and convergence by densifying sparse LiDAR observations, while LiV-GS \cite{xiao2024liv} incorporates LiDAR-vision constraints for outdoor Gaussian-splatting SLAM. Beyond initialization and densification, GS-SDF \cite{liu2025gs} couples Gaussian splatting with a neural SDF to strengthen geometric consistency, albeit with heavier modeling and optimization. Despite this progress, most LiDAR-assisted 3DGS pipelines still learn Gaussian parameters primarily from RGB photometric gradients, with LiDAR geometry rarely imposed as a persistent constraint, making them brittle under extreme illumination changes and weak texture \cite{hong2024liv,xie2024gs}. In contrast, LIT-GS combines illumination-robust thermal supervision \cite{chen2024thermal3d,lu2024thermalgaussian} with persistent LiDAR plane consistency, injecting geometric constraints into both pose/structure refinement and differentiable Gaussian optimization, thus offering a lightweight yet geometry-aware route tailored to low-contrast thermal data.

\section{METHODOLOGY}
The overall architecture of LIT-GS is illustrated in Fig.~\ref{fig:system_overview}, where solid arrows denote operation flow and orange arrows indicate gradient back-propagation. 
Hardware synchronization is achieved via PPS triggers from a microcontroller, ensuring consistent timestamp alignment among the LiDAR, IMU, and thermal camera (Fig.~\ref{fig:eqiup} (a3)).

LIT-GS is formulated as a unified geometry-aware optimization framework. 
Given synchronized LiDAR, inertial, and thermal measurements, it jointly estimates camera poses, 3D structure, and Gaussian parameters by minimizing a differentiable objective that couples thermal photometric residuals with LiDAR-derived geometric residuals. 
This formulation supports stable cross-modal estimation under weak thermal gradients and large illumination variations. LIT-GS integrates three tightly coupled components:
\begin{itemize}
    \item A confidence-aware cross-modal anchoring module that uses uncertainty-tagged visual map points from an upstream FAST-LIVO2 LiDAR-inertial-visual estimator as geometric anchors to establish thermal-LiDAR correspondences and weight residuals.
    \item A LiDAR-plane-constrained refinement stage that incorporates weighted point-to-plane residuals into joint pose/structure optimization by extending COLMAP-PCD bundle adjustment with anchor-aware weighting and adaptive residual filtering.
    \item A geometry-regularized Gaussian optimization objective that enforces local LiDAR plane consistency during differentiable splatting, compensating for weak supervision in thermally homogeneous regions.
\end{itemize}

By injecting LiDAR geometric priors into both refinement and Gaussian learning, LIT-GS reduces drift and thickness artifacts and yields a globally consistent Gaussian radiance field. 
The representation evolves from an initial LiDAR voxel map to a refined multimodal structure and finally to a compact Gaussian map for real-time rendering.

\subsection{Preprocessing}

Prior to processing, PPS-based hardware synchronization is applied and the thermal camera-LiDAR intrinsics/extrinsics are calibrated.

\subsubsection{Thermal image preprocessing}
Long-wave thermal images exhibit low contrast and weak texture, so CLAHE is applied to enhance gradients and improve keypoint detectability.
To avoid instability caused by the thermal sensor’s internal non-uniformity correction (NUC), consecutive duplicate frames are detected and skipped during visual updates.

\subsubsection{Thermal feature extraction and matching}

For frame-to-frame registration and scene-graph construction, we employ SuperPoint~\cite{detone2018superpoint} for keypoint detection and description and SuperGlue~\cite{sarlin2020superglue} for context-aware matching. Although trained on RGB imagery, these models respond to structural gradients and corner-like patterns rather than color, making them suitable for grayscale thermal images. Candidate correspondences are filtered using mutual checks and epipolar-geometry verification before backend optimization.

\subsubsection{Separation of roles: anchors vs.\ non-anchor points}

Visual map points from an upstream LiDAR-inertial-visual estimator are treated as uncertainty-weighted cross-modal anchors rather than generic SfM correspondences. 
Due to their tight coupling with LiDAR geometry, these anchors typically exhibit stronger multi-view consistency and are therefore assigned higher weights in the subsequent LiDAR-plane-constrained bundle adjustment.

SuperPoint+SuperGlue matches generate additional non-anchor points that complement anchors by improving spatial coverage and graph connectivity, especially in thermally homogeneous regions, but may contain higher noise. Accordingly, anchors dominate geometric constraints, while non-anchor points are down-weighted and filtered through adaptive residual rejection, balancing geometric stability and correspondence coverage.

\subsection{Data Refinement}

Due to residual errors in the thermal-LiDAR extrinsic calibration, preprocessing provides only coarse cross-modal alignment and may introduce structural inconsistencies. 
In addition, image-based correspondences often deteriorate under rapid motion because of motion blur and reduced tracking stability. 
To improve global geometric accuracy and robustness in dynamic scenarios, we perform a LiDAR-plane-constrained bundle adjustment (BA) that jointly refines camera poses and triangulated 3D points with frame-wise, anchor-aware geometric weighting.

\subsubsection{Frame-wise anchor-aware geometric weighting.}

To improve robustness under motion, we introduce frame-wise anchor–non-anchor geometric weighting.
For each frame $t$, triangulated 3D points are partitioned into FAST-LIVO2 anchors $\mathcal{A}_t$ and non-anchor points $\mathcal{S}_t$ obtained from SuperPoint+SuperGlue matches \cite{zheng2024fast}. Anchor geometric weight is quantified using the posterior covariance $\Sigma_k$ associated with each anchored point:
\begin{equation}
w'_k=\frac{1}{\sqrt{\det(\Sigma_k)}} .
\end{equation}
To adapt the anchor/non-anchor balance to the motion state of the current frame, we compute a normalized motion score from the linear and angular speeds using dataset-level statistics, and map it to a frame-wise anchor ratio $\alpha_t \in [\alpha_{\min}, \alpha_{\max}]$ via a sigmoid function.

In each frame, anchors are enforced to contribute a fraction $\alpha_t$ of the total geometric weight, while the remaining weight is distributed to non-anchor points according to their spatial proximity to the nearest anchor within the same frame. For non-anchor points, the geometric weight is denoted as $\tilde{w}_{k,t}$. The unified weight for both anchor and non-anchor points is defined as $w_k = w'_k$ if $\mathbf{X}_k \in \mathcal{A}_t$, and $w_k = \tilde{w}_{k,t}$ if $\mathbf{X}_k \in \mathcal{S}_t$.
The resulting weights $w_k$ are then propagated into the subsequent LiDAR-plane constrained bundle adjustment.

In all experiments, we set $\alpha_{\min}=0.5$ and $\alpha_{\max}=0.7$.
\subsubsection{LiDAR-plane constrained bundle adjustment.}

The geometric objective is defined over the filtered set $\mathcal{P}$ as
\begin{align}
\mathbf{E}_d =
\sum_{\mathbf{X}_k \in \mathcal{P}}
w_k
\left\| n_q^\top (\mathbf{X}_k - l_q) \right\|,
\end{align}
where $(l_q, n_q)$ denotes the associated LiDAR point and its local plane normal. 
This geometric term is jointly minimized with reprojection errors in a unified BA framework.

After iterative local and global BA refinement, the system produces geometrically consistent 3D structure and optimized camera poses, effectively reducing scale drift and structural misalignment.

\subsection{ Rendering Optimization}

\subsubsection{Initialization of Gaussians}

Each Gaussian primitive is defined as 
$g_i=\{\alpha_i, \mathbf{s}_i, \boldsymbol{\mu}_i, \boldsymbol{\Sigma}_i\}$,
where $\alpha_i$ denotes opacity, $\mathbf{s}_i$ the spherical harmonic (SH) coefficients, $\boldsymbol{\mu}_i$ the Gaussian center, and $\boldsymbol{\Sigma}_i$ its anisotropic covariance.

The refined 3D structure contains anchored map points, some of which provide covariance estimates that encode local geometric uncertainty. 
We therefore initialize anisotropic Gaussians from the scene-graph pair containing the largest number of such covariance-aware points, instead of using isotropic primitives. 
These points are projected onto the corresponding image to initialize the DC intensity component via bilinear interpolation, providing stable initialization for subsequent Gaussian optimization.

\subsubsection{Spherical Harmonic Coefficient Optimization}

Thermal images are grayscale, where pixel intensity reflects temperature variations. 
To model view-dependent thermal radiance, we employ third-order real spherical harmonics (SH). 
The radiance of the $k$-th Gaussian along viewing direction $\mathbf{v}$ is defined as
\begin{align}
c_k(\mathbf{v}) = \mathbf{s}_k^\top Y(\mathbf{v}),
\end{align}
where $Y(\mathbf{v})$ denotes the SH basis and $\mathbf{s}_k$ the corresponding coefficients.

Following standard alpha compositing, the accumulated transmittance before Gaussian $k$ is
\begin{align}
T_k(u_i) = \prod_{j<k} \left(1 - \alpha_j G_j(u_i)\right),
\end{align}
where $G_j(u_i)$ represents projected 2D Gaussian contribution at pixel $u_i$. 
The rendered pixel intensity is computed as
\begin{align}
\hat{C}(u_i) = \sum_{k} T_k(u_i)\, \alpha_k\, G_k(u_i)\, c_k(\mathbf{v}).
\end{align}

The photometric objective is defined as
\begin{align}
Loss_{\mathrm{pho}} = \sum_{i=1}^{M} 
\left\| C(u_i) - \hat{C}(u_i) \right\|_2^2,
\end{align}
and all Gaussian parameters, including $\mathbf{s}_k$, $\alpha_k$, $\boldsymbol{\mu}_k$, and $\boldsymbol{\Sigma}_k$, are optimized through differentiable rendering across all views.

\subsubsection{Geometric Structure Optimization}

To constrain the geometric structure of the Gaussian map, we introduce a LiDAR-plane regularization term that penalizes the point-to-plane distance between rendered 3D points and LiDAR-derived local planes.

For each pixel $u_i=(u,v)$, the rendered depth is computed as the weighted average of Gaussian depths along the corresponding camera ray:
\begin{align}
D_{\text{render}}(u_i) = \frac{\sum_k \omega_k(u_i)\, z_k}{\sum_k \omega_k(u_i)},
\end{align}
where $\omega_k(u_i)$ denotes the alpha-compositing weight of Gaussian $k$ and $z_k$ is its depth along the ray.  
The rendered 3D point is obtained by back-projection
\begin{align}
P_{\text{render}}(u,v) =
D_{\text{render}}(u,v)\, K^{-1}
\begin{bmatrix}
u \\ v \\ 1
\end{bmatrix},
\end{align}
where $K$ is the intrinsic matrix.

LiDAR points projected onto the image provide local plane supervision. 
Let $P_i$ be the LiDAR point associated with pixel $u_i$ and $n_i$ its estimated surface normal. 
The geometric loss is defined as the point-to-plane residual
\begin{align}
Loss_{\text{geo}} = \sum_i \left\| n_i^\top (P_i - P_{\text{render},i}) \right\|^2 .
\end{align}

The final objective jointly optimizes photometric and geometric terms:
\begin{align}
\mathcal{L} = (1-\lambda)Loss_{\text{pho}} + \lambda Loss_{\text{geo}},
\end{align}
where $\lambda$ balances photometric supervision and LiDAR-plane regularization. 
In our implementation, $\lambda=0.6$, slightly emphasizing geometric constraints to improve structural consistency under weak thermal gradients.

\subsubsection{Adaptive Gaussian Representation Control}\label{optimization}

After optimization stabilizes, adaptive control is applied to regulate the spatial distribution and complexity of Gaussian primitives while maintaining geometric fidelity. In geometrically complex or LiDAR-dense regions, Gaussians are densified to capture fine structures, whereas redundant primitives in simple or weakly constrained areas are pruned or merged to improve efficiency. With LiDAR plane priors, Gaussians that significantly deviate from their associated local planes are treated as geometric outliers and removed. Additionally, primitives with persistently low opacity or strong spatial redundancy are pruned, while regions with large reconstruction residuals trigger Gaussian splitting to increase local expressiveness.

\begin{figure*}[t]
    \begin{center}
        {\includegraphics[width=2.0\columnwidth]{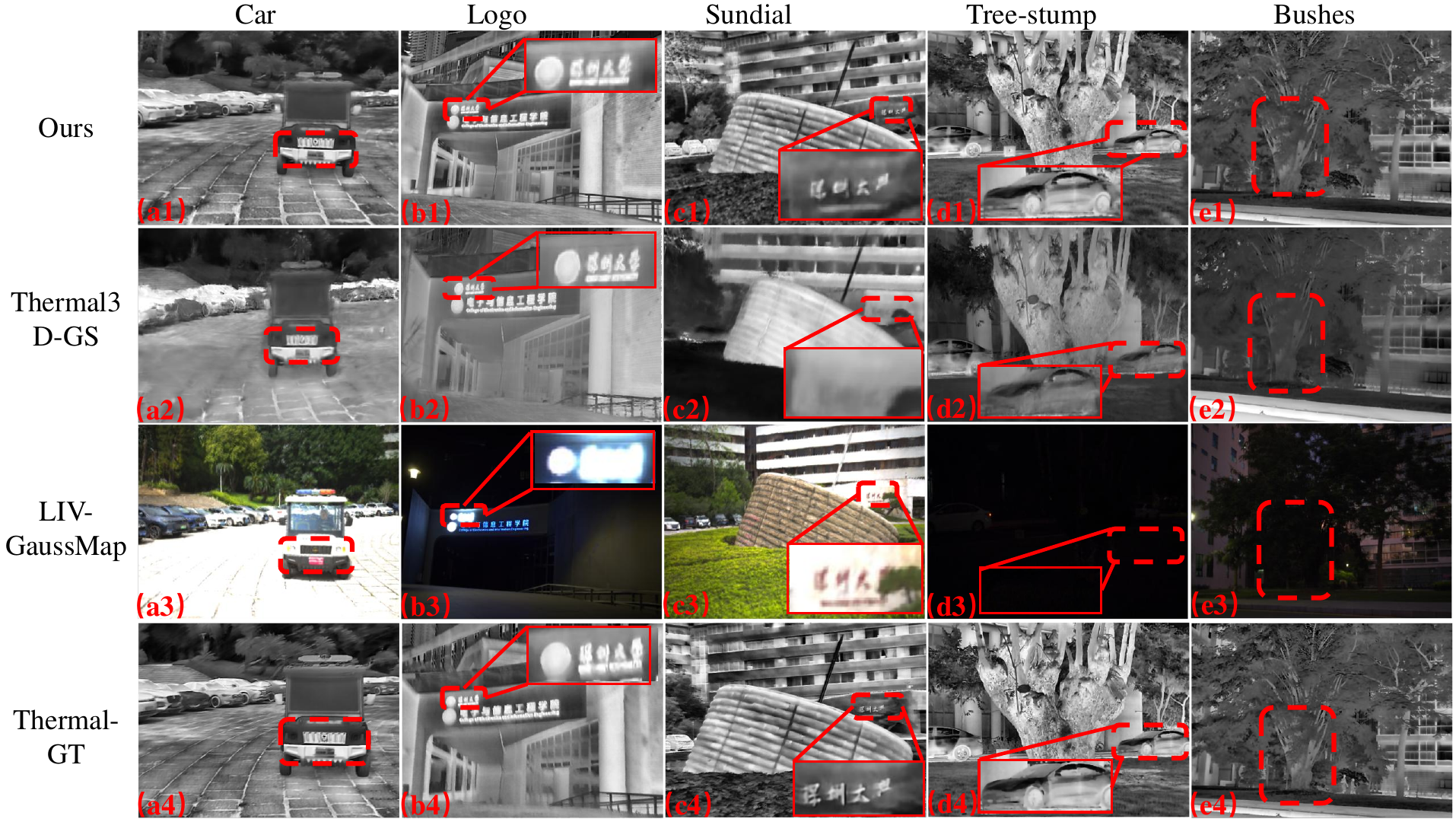}}
    \end{center}
    \vspace{-0.5cm}
    \caption{Comparison of renderings on private datasets. Data collection times (left to right) are 12:00 p.m., 10:00 p.m., 2:00 p.m., 7:00 p.m., and 6:00 a.m.}
    \label{priv_note} 
    \vspace{-0.6cm}
\end{figure*}
\subsection{Map Management}

LIT-GS adopts a unified map abstraction to organize multimodal inputs and intermediate states throughout the pipeline. The map stores several key entities, including frames (timestamps and synchronization metadata), camera poses, visual landmarks with track observations, LiDAR references for geometric constraints, and a scene graph describing image correspondences. These entities serve as a shared data interface across different processing stages: preprocessing inserts initial poses, map points, and feature correspondences, while subsequent modules update refined poses, optimized 3D points, and the final Gaussian representation, enabling modular data access and consistent state management.

\section{EXPERIMENTS AND RESULTS}
To comprehensively evaluate the proposed system, we compare it with two representative Gaussian splatting baselines: an RGB-LiDAR Gaussian mapping system (LIV-GaussMap) and a thermal-based Gaussian splatting method (Thermal3D-GS), on both our self-collected datasets and publicly available benchmarks. 
The results consistently show that the proposed method achieves improved structural integrity and rendering quality under challenging low-light conditions.

\subsection{Implementation Details}  


Data acquisition is performed using a modified version of the sensor suite\footnote{\url{https://github.com/xuankuzcr/LIV_handhold}}. 
The system consists of a Livox Avia LiDAR (with a built-in IMU), an MV-CA013-21UC visible-light camera, and an MV-CI003-GL-N15 long-wave thermal imager, which are synchronized in both time and space. 
All algorithms run on a workstation equipped with an Intel Core i7-14700KF CPU and an NVIDIA GeForce RTX 4080D GPU.
Thermal-camera intrinsics are calibrated using an asymmetric circular target composed of two materials. 
At a distance of about 1\,m, the target occupies roughly half of the thermal field of view. 
Forty images are collected by tilting the target and varying its position and orientation across the field of view, and the intrinsic parameters are estimated using Zhang’s calibration method~\cite{zhang2002flexible}. 
For extrinsic calibration, FAST-Calib~\cite{zheng2026fast} and MFCalib~\cite{ye2024mfcalib} are applied to estimate the rigid transformation between the LiDAR and the thermal camera (Fig.~\ref{fig:eqiup}(a1-a2)). 
Experiments are conducted on five self-collected datasets acquired using the proposed sensor platform, as well as five public sequences from the M2DGR benchmark~\cite{yin2021m2dgr}.

\begin{figure*}[htbp]
    \begin{center}
        {\includegraphics[width=2.05\columnwidth]{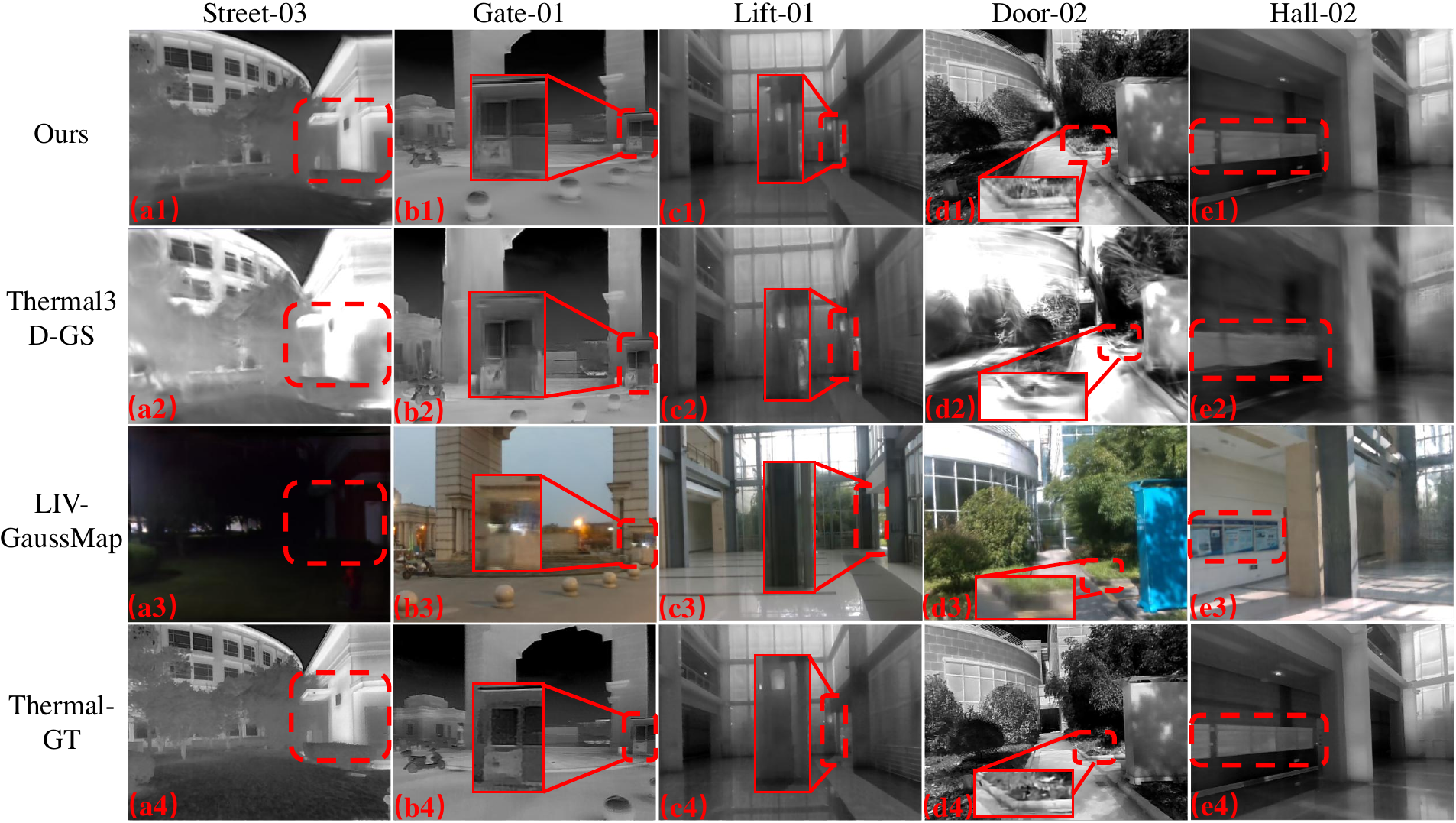}}
    \end{center}
    \vspace{-0.5cm}
    \caption{Comparison of renderings on public datasets.}
    \label{pub_note} 
    \vspace{-0.5cm}
\end{figure*}

\subsection{Comparative Experiments}

As shown in Fig.~\ref{priv_note} and Fig.~\ref{pub_note}, we compare the proposed LIT-GS with two representative Gaussian splatting baselines: LIV-GaussMap, an RGB-LiDAR Gaussian mapping system, and Thermal3D-GS, a thermal-based Gaussian splatting method. 
For each scene, the rows correspond to our method (\textit{Ours}), Thermal3D-GS, LIV-GaussMap, and the thermal observation from the same viewpoint (\textit{Thermal-GT}). 
Red boxes highlight representative regions where reconstruction artifacts are particularly visible. 
Table~\ref{tab:comparison} reports quantitative results on both private and public datasets, while Table~\ref{tab:ablation} evaluates the effect of the proposed refinement module. 
Colored cells indicate the best-performing method for each metric.

\subsubsection{Qualitative Analysis}

Across sequences captured under different lighting conditions, RGB-based Gaussian mapping often suffers from unstable photometric supervision caused by overexposure, glare, or insufficient illumination. 
This limitation can be clearly observed in Fig.~\ref{priv_note}. 
For example, in the \textit{Car} scene, the RGB-driven reconstruction produced by LIV-GaussMap loses structural details around the vehicle boundary in poorly illuminated areas, while LIT-GS preserves clearer geometric contours. 
In the \textit{Logo} and \textit{Sundial} scenes, unstable photometric gradients lead to blurred characters and distorted surface structures in LIV-GaussMap. 
Similarly, in the \textit{Tree-stump} scene, the reconstructed trunk surfaces appear noticeably thicker, indicating that weak photometric constraints allow Gaussian primitives to expand along geometrically ambiguous directions.

Thermal3D-GS enhances robustness to illumination variations by optimizing Gaussian primitives directly from thermal observations, which are largely invariant to changes in visible lighting. 
This behavior can be observed in Fig.~\ref{pub_note}, where thermal-based reconstruction preserves the overall structure of outdoor scenes even under low illumination, such as in \textit{Street-03}. 
However, because Thermal3D-GS relies solely on photometric supervision without explicit geometric regularization, its reconstructions often exhibit blurred edges and locally distorted surfaces, particularly in thermally homogeneous regions where gradients are weak. 
For instance, in the \textit{Gate-01} and \textit{Lift-01} scenes, planar structures such as walls and doors appear overly smoothed, and object boundaries are less well defined compared with those reconstructed by LIT-GS.

In contrast, LIT-GS integrates illumination-robust thermal observations with LiDAR-derived plane constraints introduced during both pose refinement and Gaussian optimization. 
This joint supervision stabilizes the reconstructed geometry and suppresses surface expansion along poorly constrained directions. 
Consequently, as highlighted in both Fig.~\ref{priv_note} and Fig.~\ref{pub_note}, LIT-GS consistently produces sharper boundaries, clearer structural details, and more geometrically consistent planar regions across all scenes, particularly in challenging environments such as nighttime, strong sunlight, and partially shadowed outdoor areas.

\begin{table*}[t]
\centering
\caption{Quantitative evaluation of our method against prior work}
\vspace{-0.3cm}
\scriptsize
\begin{tabularx}{\textwidth}{c c *{12}{c}}
\toprule
& & \multicolumn{6}{c}{\textbf{Private Datasets}} & \multicolumn{6}{c}{\textbf{Public Datasets}} \\
\cmidrule(lr){3-8} \cmidrule(lr){9-14}
\textbf{Method} & \textbf{Metric} & \textbf{Car} & \textbf{Logo} & \textbf{Sundial} & \textbf{Tree-stump} & \textbf{Bushes} & \textbf{Avg.} & \textbf{Street-03} & \textbf{Gate-01} & \textbf{Lift-01} & \textbf{Door-02} & \textbf{Hall-02} & \textbf{Avg.} \\ 
\midrule

\multirow{4}{*}{\textbf{Ours}} 
& PSNR  & \cellcolor{blue!20}{26.061} & 28.040 & 25.717 & 26.565 & 27.356 & 26.748 & 27.127 & \cellcolor{blue!20}{32.677} & \cellcolor{blue!20}{35.548} & \cellcolor{blue!20}{23.057} & \cellcolor{blue!20}{19.750} & \cellcolor{blue!20}{27.632} \\
& SSIM  & \cellcolor{green!20}{0.815} & \cellcolor{green!20}{0.833} & 0.775 & 0.845 & 0.793 & 0.812 & 0.813 & 0.903 & \cellcolor{green!20}{0.958} & 0.735 & 0.714 & \cellcolor{green!20}{0.825} \\
& LPIPS & 0.280 & 0.323 & \cellcolor{red!20}{0.266} & 0.254 & \cellcolor{red!20}{0.232} & \cellcolor{red!20}{0.271} & \cellcolor{red!20}{0.272} & \cellcolor{red!20}{0.372} & 0.193 & 0.383 & \cellcolor{red!20}{0.356} & \cellcolor{red!20}{0.315} \\
& EMD   & \cellcolor{yellow!30}{0.157} & \cellcolor{yellow!30}{0.235} & 0.213 & 0.209 & 0.140 & \cellcolor{yellow!30}{0.191} & \cellcolor{yellow!30}{0.137} & \cellcolor{yellow!30}{0.077} & \cellcolor{yellow!30}{0.087} & 0.163 & \cellcolor{yellow!30}{0.098} & \cellcolor{yellow!30}{0.112} \\

\addlinespace

\multirow{4}{*}{\textbf{Thermal3D-GS}} 
& PSNR  & 24.216 & 30.152 & \cellcolor{blue!20}{26.927} & 28.736 & 24.243 & 26.855 & 21.271 & 23.042 & 28.321 & 18.863 & 12.838 & 20.867 \\
& SSIM  & 0.808 & 0.819 & \cellcolor{green!20}{0.882} & 0.803 & \cellcolor{green!20}{0.824} & \cellcolor{green!20}{0.827} & 0.744 & 0.778 & 0.918 & 0.626 & 0.796 & 0.773 \\
& LPIPS & 0.282 & \cellcolor{red!20}{0.220} & 0.334 & \cellcolor{red!20}{0.205} & 0.390 & 0.286 & 0.511 & 0.496 & 0.385 & 0.570 & 0.514 & 0.495 \\
& EMD   & 0.227 & 0.297 & \cellcolor{yellow!30}{0.090} & \cellcolor{yellow!30}{0.179} & \cellcolor{yellow!30}{0.103} & 0.179 & 0.148 & 0.187 & 0.183 & 0.231 & 0.194 & 0.189 \\

\addlinespace

\multirow{4}{*}{\textbf{LIV-GaussMap}} 
& PSNR  & 23.569 & \cellcolor{blue!20}{31.355} & 22.142 & \cellcolor{blue!20}{32.204} & \cellcolor{blue!20}{30.051} & \cellcolor{blue!20}{27.864} & \cellcolor{blue!20}{29.222} & 25.908 & 28.081 & 18.340 & 17.303 & 24.711 \\
& SSIM  & \cellcolor{green!20}{0.815} & 0.709 & 0.650 & \cellcolor{green!20}{0.888} & \cellcolor{green!20}{0.810} & 0.774 & \cellcolor{green!20}{0.865} & 0.760 & 0.893 & 0.725 & 0.685 & 0.786 \\
& LPIPS & \cellcolor{red!20}{0.181} & 0.372 & 0.303 & 0.338 & 0.302 & 0.299 & 0.332 & 0.417 & \cellcolor{red!20}{0.181} & \cellcolor{red!20}{0.322} & 0.412 & 0.333 \\
& EMD   & 0.280 & 0.247 & 0.297 & 0.314 & 0.298 & 0.287 & 0.163 & 0.250 & 0.202 & \cellcolor{yellow!30}{0.121} & 0.383 & 0.224 \\

\bottomrule
\end{tabularx}
\label{tab:comparison}
\vspace{-0.1cm}
\end{table*}

\begin{table*}[ht]
\centering
\caption{Ablation Study of Data Refinement}
\vspace{-0.3cm}
\scriptsize
\begin{tabularx}{\textwidth}{c c *{12}{c}}
\toprule
& & \multicolumn{6}{c}{\textbf{Private Datasets}} & \multicolumn{6}{c}{\textbf{Public Datasets}} \\
\cmidrule(lr){3-8} \cmidrule(lr){9-14}
\textbf{Method} & \textbf{Metric} & \textbf{Car} & \textbf{Logo} & \textbf{Sundial} & \textbf{Tree-stump} & \textbf{Bushes} & \textbf{Avg.} & \textbf{Street-03} & \textbf{Gate-01} & \textbf{Lift-01} & \textbf{Door-02} & \textbf{Hall-02} & \textbf{Avg.} \\ 
\midrule
\multirow{4}{*}{\textbf{w/ refine}} 
& PSNR   & \cellcolor{blue!20}26.061 & \cellcolor{blue!20}28.040 & \cellcolor{blue!20}25.717 & \cellcolor{blue!20}26.565 & \cellcolor{blue!20}27.356 & \cellcolor{blue!20}26.748 & \cellcolor{blue!20}27.127 & \cellcolor{blue!20}32.677 & \cellcolor{blue!20}35.548 & \cellcolor{blue!20}23.057 & \cellcolor{blue!20}19.750 & \cellcolor{blue!20}27.632 \\ 
& SSIM   & \cellcolor{green!20}0.815 & \cellcolor{green!20}0.833 & \cellcolor{green!20}0.775 & \cellcolor{green!20}0.845 & \cellcolor{green!20}0.793 & \cellcolor{green!20}0.812 & \cellcolor{green!20}0.813 & \cellcolor{green!20}0.903 & \cellcolor{green!20}0.958 & \cellcolor{green!20}0.735 & \cellcolor{green!20}0.714 & \cellcolor{green!20}0.825 \\ 
& LPIPS  & \cellcolor{red!20}0.280 & \cellcolor{red!20}0.323 & \cellcolor{red!20}0.266 & \cellcolor{red!20}0.254 & \cellcolor{red!20}0.232 & \cellcolor{red!20}0.271 & \cellcolor{red!20}0.272 & \cellcolor{red!20}0.372 & \cellcolor{red!20}0.193 & \cellcolor{red!20}0.383 & \cellcolor{red!20}0.356 & \cellcolor{red!20}0.315 \\ 
& EMD    & \cellcolor{yellow!30}0.157 & \cellcolor{yellow!30}0.235 & \cellcolor{yellow!30}0.213 & \cellcolor{yellow!30}0.209 & \cellcolor{yellow!30}0.140 & \cellcolor{yellow!30}0.191 & \cellcolor{yellow!30}0.137 & \cellcolor{yellow!30}0.077 & \cellcolor{yellow!30}0.087 & \cellcolor{yellow!30}0.163 & \cellcolor{yellow!30}0.098 & \cellcolor{yellow!30}0.112 \\ 
\addlinespace
\multirow{4}{*}{\textbf{w/o refine}} 
& PSNR   & 21.038 & 21.156 & 23.702 & 24.491 & 22.084 & 22.494 & 18.901 & 23.233 & 31.013 & 18.781 & 17.239 & 21.833 \\ 
& SSIM   & 0.594 & 0.705 & 0.735 & 0.818 & 0.708 & 0.712 & 0.695 & 0.779 & 0.938 & 0.624 & 0.693 & 0.746 \\ 
& LPIPS  & 0.289 & 0.377 & 0.296 & 0.291 & 0.330 & 0.317 & 0.536 & 0.487 & 0.338 & 0.568 & 0.509 & 0.488 \\ 
& EMD    & 0.228 & 0.236 & 0.308 & 0.228 & 0.345 & 0.269 & 0.214 & 0.265 & 0.276 & 0.169 & 0.289 & 0.243 \\ 
\bottomrule
\end{tabularx}
\label{tab:ablation}
\vspace{-0.15cm}
\end{table*}
\subsubsection{Quantitative Analysis}

Reconstruction quality is evaluated using geometric and rendering metrics. 
Earth Mover’s Distance (EMD) measures geometric consistency between reconstructed and reference point clouds, while PSNR, SSIM, and LPIPS evaluate rendering fidelity. 
Since PSNR and SSIM may be artificially high in extremely dark scenes, we interpret them jointly with LPIPS and the geometry metric EMD.

As shown in Table~\ref{tab:comparison}, LIT-GS consistently achieves the best overall performance across most datasets. 
Compared with LIV-GaussMap, LIT-GS significantly reduces EMD across both private and public scenes, indicating improved geometric consistency. 
The improvements are particularly evident in scenes such as \textit{Gate-01} and \textit{Lift-01}, where large planar structures dominate the environment and benefit from the proposed LiDAR-plane regularization. 
In addition, LIT-GS achieves higher SSIM and lower LPIPS values, indicating improved perceptual rendering quality and sharper structural boundaries.

Although Thermal3D-GS performs competitively in some photometric metrics due to its illumination-invariant thermal supervision, its EMD values are generally higher than those of LIT-GS, reflecting weaker structural consistency. 
This observation suggests that purely thermal photometric optimization is insufficient to ensure reliable geometric reconstruction in regions with weak temperature gradients. 
The results highlight the importance of integrating explicit geometric constraints when optimizing Gaussian primitives. 
Furthermore, the ablation results in Table~\ref{tab:ablation} show that disabling the proposed refinement module leads to clear degradation in both geometric and rendering metrics, confirming that the LiDAR-plane-constrained refinement plays a critical role in stabilizing pose estimation and improving the final Gaussian reconstruction.

\begin{figure*}[htbp]
    \begin{center}
        {\includegraphics[width=1.9\columnwidth]{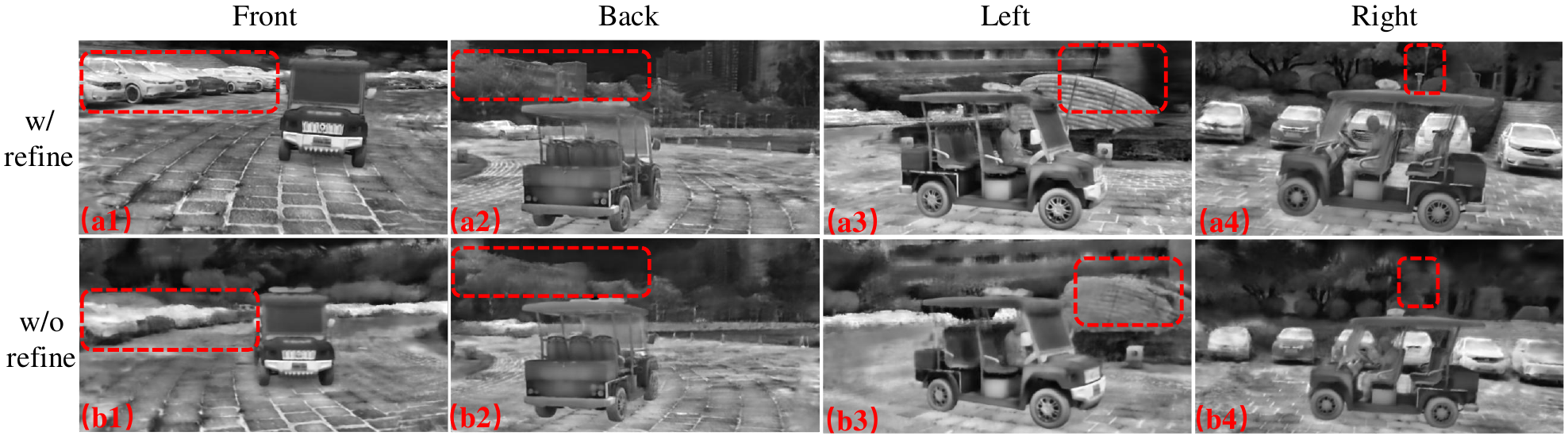}}
    \end{center}
    \vspace{-0.5cm}
    \caption{ Omni-view ablation experiment demonstration. From the perspectives of front, back, left, and right, the refined perspectives (a1-a4) are compared with the unrefined perspectives (b1-b4). The unrefined perspective structure is not well constrained.}
    \label{ablt:omni-view} 
    \vspace{-0.6cm}
\end{figure*}

\subsection{Runtime and Computational Cost Analysis}

Training on the \textit{Car} scene takes 70.6 minutes.
After preprocessing and data refinement, the average rendering time for a single frame at a resolution of $640\times512$ is approximately 25.3 ms. 
Although our method introduces additional geometry-aware constraints during optimization, the explicit Gaussian representation together with splatting-based rasterization remains computationally efficient. 

\subsection{Ablation Study}

\subsubsection{Evaluation of Learned Feature Matching under Low-Contrast Thermal Imagery}

To evaluate feature robustness in low-contrast thermal scenes, we compare ORB, SIFT, SP+SG (SuperPoint+SuperGlue), and SP+SG+CLAHE in the front-end while keeping all other settings fixed. 
As shown in Table~\ref{tab:thermal_ablation}, learned features (SP+SG) significantly increase the inlier ratio and reduce both ATE and LiDAR-plane residuals compared with hand-crafted methods, indicating stronger geometric consistency and more reliable multi-view associations. 
Applying CLAHE further improves matching stability and achieves the best overall performance in most scenes. 
Since this ablation focuses on front-end feature robustness, we mainly report geometric metrics (inlier ratio, ATE, and plane residual), while rendering metrics are evaluated in the refinement ablation presented in Table~\ref{tab:ablation}.

\subsubsection{Ablation Study of Data Refinement}

Table~\ref{tab:ablation} evaluates the proposed data refinement strategy under identical training settings. 
Introducing refinement consistently improves PSNR and SSIM while reducing LPIPS and EMD across both private and public datasets, indicating improved photometric fidelity and structural consistency. 
Qualitative results in Fig.~\ref{ablt:omni-view} further show clearer object boundaries and more stable global structures, demonstrating that the geometry-aware refinement effectively stabilizes pose estimation and suppresses structural drift during Gaussian optimization.

\begin{table*}[ht!]
\centering
\caption{Ablation Study on Thermal Feature Extraction and Matching}
\vspace{-0.3cm}
\scriptsize
\begin{tabularx}{\textwidth}{c c *{12}{c}}
\toprule
& & \multicolumn{6}{c}{\textbf{Private Datasets}} & \multicolumn{6}{c}{\textbf{Public Datasets}} \\
\cmidrule(lr){3-8} \cmidrule(lr){9-14}
\textbf{Method} & \textbf{Metric} 
& \textbf{Car} & \textbf{Logo} & \textbf{Sundial} & \textbf{Tree-stump} & \textbf{Bushes} & \textbf{Avg.} 
& \textbf{Street-03} & \textbf{Gate-01} & \textbf{Lift-01} & \textbf{Door-02} & \textbf{Hall-02} & \textbf{Avg.} \\ 
\midrule

\multirow{4}{*}{\textbf{ORB}} 
& Inlier (\%)        & 41.8 & 46.2 & 47.5 & 50.6 & 44.1 & 46.0 & 38.9 & 46.9 & 53.6 & 39.2 & 37.4 & 43.2 \\
& ATE (m)            & 0.195 & 0.173 & 0.168 & 0.154 & 0.186 & 0.175 & 0.218 & 0.172 & 0.145 & 0.233 & 0.248 & 0.203 \\
& Plane Res. (m)     & 0.075 & 0.068 & 0.063 & 0.056 & 0.073 & 0.067 & 0.086 & 0.060 & 0.048 & 0.093 & 0.101 & 0.078 \\
& PSNR               & 22.100 & 23.400 & 23.800 & 24.400 & 23.200 & 23.380 & 21.600 & 24.900 & 30.100 & 20.500 & 19.300 & 23.280 \\
\addlinespace

\multirow{4}{*}{\textbf{SIFT}} 
& Inlier (\%)        & 49.1 & 52.4 & 53.0 & 55.8 & 50.2 & 52.1 & 44.7 & 54.9 & 59.4 & 44.8 & 42.9 & 49.3 \\
& ATE (m)            & 0.160 & 0.146 & 0.142 & 0.133 & 0.158 & 0.148 & 0.187 & 0.145 & 0.128 & 0.192 & 0.207 & 0.172 \\
& Plane Res. (m)     & 0.062 & 0.057 & 0.053 & 0.050 & 0.061 & 0.057 & 0.073 & 0.050 & 0.042 & 0.079 & 0.085 & 0.066 \\
& PSNR               & 23.900 & 24.900 & 25.100 & 25.600 & 24.700 & 24.840 & 23.200 & 26.900 & 31.800 & 22.000 & \cellcolor{blue!20}21.000 & 24.980 \\
\addlinespace

\multirow{4}{*}{\textbf{SP+SG}} 
& Inlier (\%)        & 63.0 & 66.2 & 61.1 & 68.4 & 63.7 & 64.5 & 59.6 & 69.8 & 71.4 & 58.1 & 56.2 & 63.0 \\
& ATE (m)            & 0.100 & 0.090 & 0.106 & 0.085 & 0.096 & 0.095 & 0.115 & 0.078 & 0.073 & 0.128 & 0.141 & 0.107 \\
& Plane Res. (m)     & 0.039 & 0.035 & 0.043 & 0.033 & 0.037 & 0.037 & 0.047 & 0.029 & 0.026 & 0.056 & 0.061 & 0.044 \\
& PSNR               & 25.000 & 26.400 & \cellcolor{blue!20}25.600 & 26.200 & 25.800 & 25.800 & 25.900 & 31.300 & 34.500 & 22.400 & 19.600 & 26.740 \\
\addlinespace

\multirow{4}{*}{\makecell{\textbf{SP+SG}\\\textbf{+CLAHE}}} 
& Inlier (\%)        & \cellcolor{green!20}70.8 & \cellcolor{green!20}69.1 & \cellcolor{green!20}66.5 & \cellcolor{green!20}73.2 & \cellcolor{green!20}69.8 & \cellcolor{green!20}69.9 
                     & \cellcolor{green!20}68.1 & \cellcolor{green!20}75.0 & \cellcolor{green!20}77.6 & \cellcolor{green!20}62.9 & \cellcolor{green!20}60.7 & \cellcolor{green!20}68.9 \\
& ATE (m)            & \cellcolor{red!20}0.078 & \cellcolor{red!20}0.084 & \cellcolor{red!20}0.098 & \cellcolor{red!20}0.072 & \cellcolor{red!20}0.080 & \cellcolor{red!20}0.082 
                     & \cellcolor{red!20}0.087 & \cellcolor{red!20}0.063 & \cellcolor{red!20}0.060 & \cellcolor{red!20}0.101 & \cellcolor{red!20}0.109 & \cellcolor{red!20}0.084 \\
& Plane Res. (m)     & \cellcolor{yellow!30}0.030 & \cellcolor{yellow!30}0.032 & \cellcolor{yellow!30}0.040 & \cellcolor{yellow!30}0.026 & \cellcolor{yellow!30}0.031 & \cellcolor{yellow!30}0.032
                     & \cellcolor{yellow!30}0.036 & \cellcolor{yellow!30}0.023 & \cellcolor{yellow!30}0.020 & \cellcolor{yellow!30}0.044 & \cellcolor{yellow!30}0.048 & \cellcolor{yellow!30}0.034 \\
& PSNR               & \cellcolor{blue!20}25.800 & \cellcolor{blue!20}27.600 & 25.400 & \cellcolor{blue!20}26.800 & \cellcolor{blue!20}26.900 & \cellcolor{blue!20}26.500
                     & \cellcolor{blue!20}26.800 & \cellcolor{blue!20}32.200 & \cellcolor{blue!20}35.000 & \cellcolor{blue!20}23.100 & 19.700 & \cellcolor{blue!20}27.360 \\
\bottomrule
\end{tabularx}
\label{tab:thermal_ablation}
\vspace{-0.5cm}
\end{table*}

\section{CONCLUSION}

This study proposes a LiDAR–inertial–thermal Gaussian Splatting approach for robust 3D reconstruction from thermal imagery. Thermal radiance provides illumination-invariant appearance supervision, while LiDAR enforces metric geometric constraints to preserve structural fidelity. High-confidence map points with uncertainty estimates from a tightly coupled LiDAR–inertial front-end are used as geometric anchors to guide cross-modal association, pose/structure refinement, and subsequent Gaussian modeling. By jointly exploiting thermal cues and LiDAR geometry, the method improves robustness in low-light and complex environments and maintains high reconstruction accuracy in dynamic scenes. Experiments show RGB–LiDAR Gaussian mapping outperforms thermal–LiDAR in good lighting, motivating RGB–thermal fusion within a LiDAR-regularized Gaussian framework for robust performance across lighting conditions.

\bibliographystyle{IEEEtran}
\balance
\bibliography{paper}

\end{document}